\documentclass[nonatbib]{article}

\usepackage[final]{neurips_2021}
\usepackage[utf8]{inputenc} 
\usepackage[T1]{fontenc}    
\usepackage[ruled,vlined]{algorithm2e}
\usepackage{algorithmic}
\usepackage{amsfonts}       
\usepackage{amsmath}
\usepackage{bm}
\usepackage{booktabs}       
\usepackage{graphicx}
\usepackage{hyperref}       
\usepackage{microtype}      
\usepackage{nicefrac}       
\usepackage{subfig}
\usepackage{url}            
\usepackage{xcolor}         

\newlength{\bibitemsep}
\newlength{\bibparskip}
\let\oldthebibliography\thebibliography
\renewcommand\thebibliography[1]{%
  \oldthebibliography{#1}%
  \setlength{\parskip}{\bibitemsep}%
  \setlength{\itemsep}{\bibparskip}%
}

\usepackage{listings}\title{Grand Challenge On Detecting Cheapfakes}
\author{
   Duc-Tien Dang-Nguyen\textsuperscript{1, 3}, Sohail Ahmed Khan\textsuperscript{1}, Cise Midoglu\textsuperscript{2}, \\
   \textbf{Michael Riegler\textsuperscript{2}, P{\aa}l Halvorsen\textsuperscript{2}, Minh-Son Dao\textsuperscript{4}} \\
   \textsuperscript{1}University of Bergen (UiB), \textsuperscript{2}Simula Metropolitan Center for Digital Engineering (SimulaMet),\\ \textsuperscript{3}Kristiania University College, \textsuperscript{4}National Institute of Information and Communications Technology (NICT)\\
   \texttt{\{ductien.dangnguyen, sohail.khan\}@uib.no, \{cise, michael, paalh\}@simula.no, dao@nict.go.jp}
}

\begin{document}

\maketitle

\vspace{-0.5cm}
\begin{figure}[ht!]
    \centering
    \includegraphics[width=0.95\linewidth]{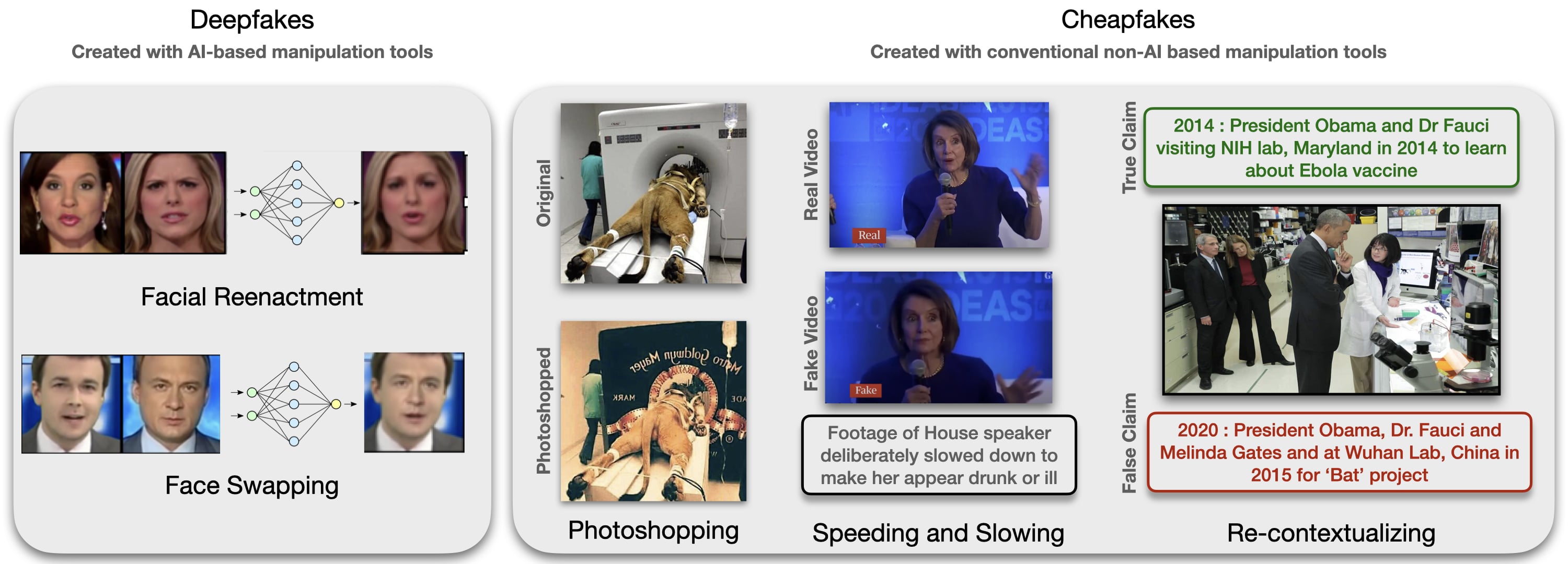}
    \caption{Deepfakes (left) are defined as falsified media created using sophisticated AI-based media manipulation tools and techniques.  Cheapfakes (right) include falsified media created with/without contemporary non-AI based editing tools which are easily accessible. Photoshopping tools can be used to tamper with images. Videos can be sped up or slowed down to change the intent or misrepresent the person in the video. Re-contextualizing includes associating falsified or unrelated claims with a genuine image to misrepresent events or persons. This challenge is focused on detecting re-contextualized cheapfakes. Image sources:~\cite{ff_dataset,photoshopped_lion,pelosi_fake,obama_maryland,obama_wuhan}}
    \label{figure:fakes-figure}
\end{figure}
\vspace{-0.5cm}


\section{Motivation and Background}

\textit{Cheapfake} is a recently coined general term that encompasses non-AI (``cheap'') manipulations of multimedia content, created without using deep learning methods. Although a lot of attention has been paid to the creation, detection, and misuse of deepfakes in the last years, cheapfakes are actually known to be more prevalent than deepfakes~\cite{factsheet-covid19, mit_tech_report}. 

Cheapfakes can be created by using contemporary editing tools, which are non-AI based and are easily accessible, such as Adobe Photoshop or PremierePro, or even without using any software. Image manipulations, speeding/slowing of videos, deliberate alteration of the context of the multimedia asset in, e.g., news captions, by sharing the media alongside misleading claims, are some of the methods that are currently in use (see Figure \ref{figure:fakes-figure}). The latter is referred to as context alteration or out-of-context (OOC) misuse of media. OOC media are much harder to detect than fake media, since the images and/or videos are not tampered. An overview of different types of cheapfakes surfacing the Internet are reported by~\cite{paris2019}.

Depending on the type of cheapfakes, different detection tools can be used. Methods to detect image manipulations such as photoshopping and image splicing have been investigated~\cite{Chen2017ImageSD, Cozzolino2015SplicebusterAN, huh2018fighting, wang2019detecting}. Re-contextualization or OOC misuse, which include associating falsified or unrelated claims with a genuine image in order to misrepresent events or persons is, however, relatively niche and unexplored. Aneja et al.~\cite{aneja2021cosmos} have recently introduced this task, provided a dataset of real-world news posts called COSMOS, and proposed a method for detecting cheapfakes, which was benchmarked using the COSMOS dataset. 

The aim of this challenge is to develop models that can be used to detect OOC images, and more specifically the misuse of real photographs with conflicting image captions in news items, based on a version of the COSMOS dataset. We have organized similar challenges on the detection of cheapfakes at the ACM Multimedia Systems Conference 2021 (MMSys'21) and the ACM Multimedia Conference 2022 (ACMMM 2022)~\cite{mmsys21-challenge-arxiv, acmmm22-challenge-arxiv}.

\begin{figure}
    \centering
    \includegraphics[width=0.7\linewidth]{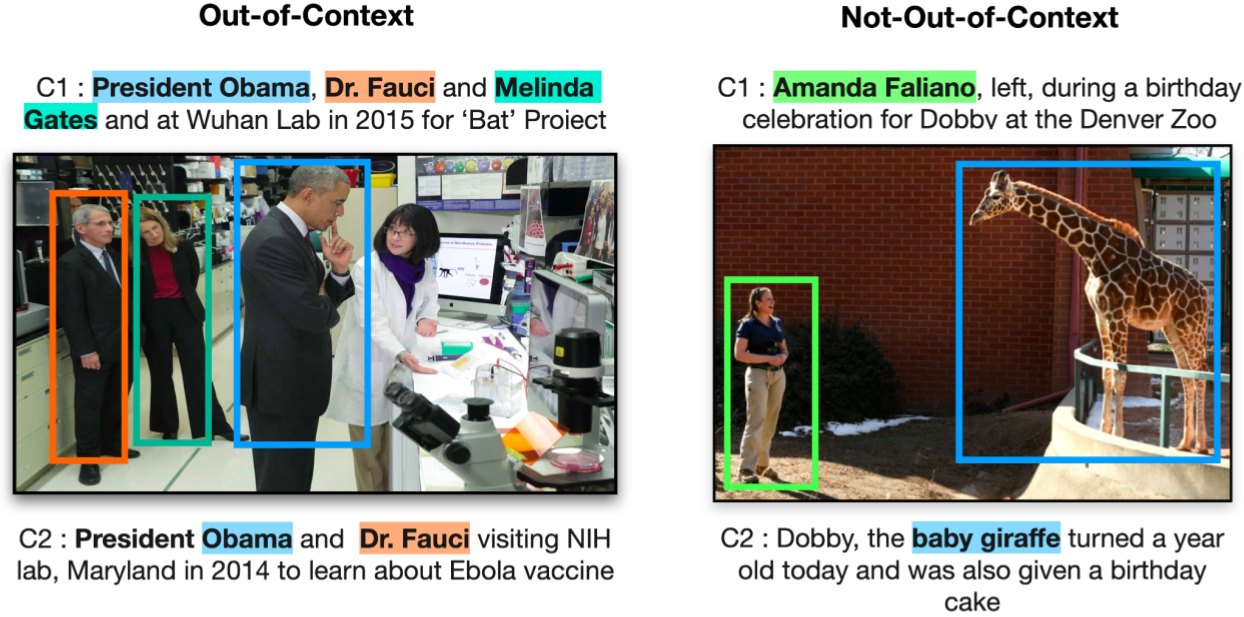}
    \caption{Each image in the COSMOS dataset is accompanied by two captions that the image was circulated together with on the Internet. Left: one of the two captions is misleading due to the alteration of context, indicating out-of-context (OOC) misuse. Right: none of the captions are misleading, hence not-out-of-context (NOOC). Image source:~\cite{aneja2021cosmos}}
    \label{figure:task-figure}
\end{figure}


\section{Host Organizations and Coordinator Contacts} 

This challenge is a collaboration between the University of Bergen (UiB) in Norway, Simula Metropolitan Center for Digital Engineering (SimulaMet) in Norway, and the National Institute of Information and Communications Technology (NICT) in Japan. Organization committee: 

\begin{itemize}
    \item Duc Tien Dang Nguyen, UiB, ductien.dangnguyen@uib.no
    \item Sohail Ahmed Khan, SFI-MediaFutures - UiB, sohail.khan@uib.no 
    \item Cise Midoglu, SimulaMet, cise@simula.no
    \item Michael Riegler, SimulaMet, michael@simula.no
    \item P{\aa}l Halvorsen, SimulaMet, paalh@simula.no 
    \item Minh-Son Dao, NICT, dao@nict.go.jp
\end{itemize}


\section{Challenge Description}

An image serves as evidence of an event described by a news caption. Presenting an image as evidence of untrue and/or unrelated events is defined as out-of-context (OOC) use of the image (see Figure~\ref{figure:task-figure}). The aim of this challenge is to develop and benchmark models that can be used to detect OOC misuse of images in news items. 

\subsection*{Task 1: Detection of Conflicting Image-Caption Triplets}

If two captions associated with an image are valid, then they should describe the same event. 
If they refer to same object(s) in the image, but are semantically different, i.e., associate the same subject to different events, this indicates OOC use of the image. However, if the captions correspond to the same event, irrespective of the object(s) they describe, this is defined as not-out-of-context (NOOC) use of the image. 

In this task, participants are asked to come up with methods to detect conflicting image-caption triplets, which indicate miscontextualization. More specifically, given \texttt{<Image,Caption1,Caption2>} triplets as input, their proposed model should predict corresponding class labels \texttt{1 (OOC)} or \texttt{0 (NOOC)}. The goal is not to identify individual captions as true/false, but rather to detect the existence of miscontextualization. Such methods are considered particularly useful for assisting fact checkers, as highlighting conflicting image-caption triplets allows them to narrow down their search space. 

\subsection*{Task 2: Detection of Fake Captions}

In a practical scenario, multiple captions might not be available for a given image, and the challenge boils down to figuring out whether an individual caption associated with an image is genuine or not. 

In this task, participants are asked to come up with methods to determine whether a given image-caption pair is genuine (real) or falsely generated (fake). More specifically, given an \texttt{<Image,Caption>} pair as input, their proposed model should predict corresponding class labels \texttt{0 (real)} or \texttt{1 (fake)}. 

We acknowledge that this is a challenging task without prior knowledge of the image origin, even for human moderators. In fact, Luo et al~\cite{luo2021newsclippings} have verified this challenge with a study on human evaluators, who were instructed not to use search engines, where the average human accuracy was only around $65\%$.


\section{Dataset}

For this challenge, an augmented version of the COSMOS dataset~\cite{aneja2021cosmos} will be used. A part if this dataset is sampled and assigned as the public dataset. The public dataset, consisting of the training, validation and public test splits, is provided openly to participants for training and testing their algorithms. The remaining part of the COSMOS dataset is augmented with new samples and modified to create the hidden test split, similar to~\cite{mmsys21-challenge-arxiv,acmmm22-challenge-arxiv}. The hidden test split is not made publicly available, and will be used by the challenge organizers to evaluate the submissions. Details of the challenge dataset are summarized in Table~\ref{table:dataset-statistics}.


\begin{table}[h!]
\small
    \centering
    \caption{Challenge dataset statistics.}
    \label{table:dataset-statistics}
    \begin{tabular}{|l|l|l|l|}
    \hline
    \textbf{Dataset Split} & \textbf{Number of Images} & \textbf{Number of Captions} & \textbf{Context Annotation} \\
    \hline
    Training & $161752$ & $360749$ & No \\
    \hline
    Validation & $41006$ & $90036$ & No \\
    \hline
    Public Test & $1000$ & $2000$ & Yes\\
    \hline
    Hidden Test & $1000$ & $2000$ & Yes\\
    \hline
    \end{tabular}
\end{table}
\vspace{-0.5cm}


\section{Evaluation Criteria}

Participant models will be evaluated and ranked according to two aggregate scores, composed of $5$ and $3$ metrics respectively.  

\begin{itemize}
    \item \textit{Effectiveness}: accuracy, precision, recall, F1-score, and Matthews correlation coefficient (MCC). Participants are asked to calculate these $5$ metrics for their model and include the values in the results section of their submission. 
    \item \textit{Efficiency}: latency, number of parameters, and model size. Participants are asked to calculate these $3$ metrics for their model and include the values in the results section of their submission. 
\end{itemize}

After the participants evaluate their own models on the public test split, they are asked to provide code and trained model weights to the organization committee, in order for their models to be evaluated on the hidden test split. Participants are allowed to submit their solutions in three alternative ways as described in Section~\ref{section:submission-guidelines}, provided that they abide by the deadlines listed in Section~\ref{section:important-dates}.


\section{Important Dates}\label{section:important-dates}

\begin{itemize}
    \item Dataset release (public training): Monday, 16 January 2023
    \item Dataset release (public test): Monday, 20 February 2023
    \item Model submission deadline: Wednesday, 29 March 2023
    \item Paper submission deadline: Monday, 17 March 2023
    \item Model evaluation results announcement: Monday, 03 April 2023
    \item Paper acceptance announcement: Monday, 24 April 2023
\end{itemize}


\section{Submission Guidelines}\label{section:submission-guidelines}

\subsection*{Docker Container}

We recommend challenge participants to submit their solutions as a Docker container, since it will make sure that we don’t get any errors resulting from software incompatibility issues or any other similar reason. In this case, we recommend them to follow the instructions given under \url{https://github.com/detecting-cheapfakes/detecting-cheapfakes-code}. 

\subsection*{Standard Python Executable}

If the participants face any difficulties in submitting their solutions as a Docker container, or if they feel more comfortable submitting their solution as a standard Python project, they can do so by following the instructions below. It would also be helpful for us if the participants use PyTorch as the main library if they would like to submit their Python projects, however, this is not compulsory.

\begin{itemize}
    \item We expect that the submitted code will be executable by a single command, for example: 
    \begin{verbatim}
    python solution.py <path to folder containing the hidden test split file 
    private_test.json>
    \end{verbatim}
    
    \item Participants should expect the same format for both the \texttt{private\_test.json} file and the \texttt{public\_test.json} file.
    
    \item To cope with any software incompatibility issues, we request the participants to provide a \texttt{requirements.txt} file along with their solutions, containing the names and specific version numbers of software packages used. This is fairly easy to do with both \textbf{Conda} (\texttt{conda list}), and \textbf{Pip} (\texttt{pip list}). We recommend that the participants use \textbf{Conda} and create a fresh environment before starting to write the code for their challenge solution. 
\end{itemize}

\subsection*{Jupyter Notebook}

Participants can also submit their solutions using Jupyter notebooks, by following the instructions below.

\begin{itemize}
    \item Participants should structure their code so that it allows us to change the input path by updating a single line in the main file, i.e., the submitted notebook should be executable after changing the \texttt{INPUT\_FOLDER} parameter from the path for the public test split file, to the path for the private test split file, as shown below:

    \begin{verbatim}
    INPUT_FOLDER = <path to folder containing the hidden test split file 
    private_test.json>
    \end{verbatim}    
    
    \item Participants should expect the same format for both the \texttt{private\_test.json} file and the \texttt{public\_test.json} file.
\end{itemize}



\section{Additional Information}

\subsection*{Experience}

The first two editions of this challenge have been organized within, (1) ACM Multimedia Systems Conference 2021 as the ``MMSys'21 Grand Challenge on Detecting Cheapfakes''~\cite{mmsys21-challenge-web,mmsys21-challenge-arxiv}, and (2) ACM Multimedia Conference 2022 as the ``ACM Multimedia Grand Challenge on Detecting Cheapfakes''~\cite{acmmm22-challenge-arxiv}. 

\subsection*{Participant Support}

A challenge website has been setup under \url{https://detecting-cheapfakes.github.io/}, and contains information on datasets, tasks, and other resources. The GitHub organization \url{https://github.com/detecting-cheapfakes/} hosts all relevant repositories. A Google Group \url{https://groups.google.com/g/grandchallenge-cheapfakes} has also been established to support prospective participants. Interested participants can find the previously asked questions and join interactive discussions on these platforms. 

\subsection*{Publicity}

We plan to promote the challenge through the following: (i) professional networks such as MediaFutures, Norwegian Artificial Intelligence Society (NAIS), and the Norwegian AI Community (NORA), (ii) email lists, Slack workspaces, and announcement boards, including VisionList and Image World, (iii) conferences where organizers of this challenge 
serve as OC members, such as ACM ICMR, ACM MMSys, Global Fact, IEEE CVPR, MediaEval, MMM, NAIS, and NTCIR. 

\subsection*{Continuation}

The challenge website and other participant resources have been setup with a commitment to be maintained at least for the next $3$ years. We would like to reiterate our commitment to sustain the challenge over the next years, as we believe that it is both specific, relevant and timely, as well as generic enough to evolve over time according to different needs. For instance, with the rising importance of synthetic data for research and training purposes, a possible future task could be the \textit{generation} of fake captions. 





\newpage
\small
\bibliographystyle{unsrt} 
\bibliography{references.bib}

\begin{thebibliography}{10}

\bibitem{ff_dataset}
A.~Rössler, D.~Cozzolino, L.~Verdoliva, C.~Riess, J.~Thies, and M.~Nießner.
\newblock {FaceForensics++}: Learning to detect manipulated facial images,
  2019.

\bibitem{photoshopped_lion}
Boredpanda.
\newblock 30 fake viral photos people believed were real, 2019.

\bibitem{pelosi_fake}
J.~Waterson.
\newblock Facebook refuses to delete fake {Pelosi} video spread by {Trump}
  supporters, 2019.

\bibitem{obama_maryland}
The~White House.
\newblock President {Obama} tours a lab at the {Vaccine Research Center} at the
  {National Institutes of Health}, 2014.

\bibitem{obama_wuhan}
D.~Evon.
\newblock Is this {Obama}, {Fauci}, and {Gates} at a {Wuhan Lab} in 2015?,
  2020.

\bibitem{factsheet-covid19}
J.~S. Brennen, F.~M. Simon, P.~N. Howard, and R.~K. Nielsen.
\newblock Types, sources, and claims of {COVID-19} misinformation, 2020.

\bibitem{mit_tech_report}
N.~Schick.
\newblock Don’t underestimate the cheapfake, 2020.

\bibitem{paris2019}
B.~Paris and J.~Donovan.
\newblock Deepfakes and cheapfakes: The manipulation of audio and visual
  evidence, 2019.

\bibitem{Chen2017ImageSD}
C.~Chen, S.~McCloskey, and J.~Yu.
\newblock Image splicing detection via camera response function analysis.
\newblock {\em 2017 IEEE Conference on Computer Vision and Pattern Recognition
  (CVPR)}, pages 1876--1885, 2017.

\bibitem{Cozzolino2015SplicebusterAN}
D.~Cozzolino, G.~Poggi, and L.~Verdoliva.
\newblock Splicebuster: A new blind image splicing detector.
\newblock {\em 2015 IEEE International Workshop on Information Forensics and
  Security (WIFS)}, pages 1--6, 2015.

\bibitem{huh2018fighting}
M.~Huh, A.~Liu, A.~Owens, and A.~A. Efros.
\newblock Fighting fake news: Image splice detection via learned
  self-consistency, 2018.

\bibitem{wang2019detecting}
S.~Wang, O.~Wang, A.~Owens, R.~Zhang, and A.~A. Efros.
\newblock Detecting photoshopped faces by scripting photoshop.
\newblock In {\em ICCV}, 2019.

\bibitem{aneja2021cosmos}
Shivangi Aneja, Chris Bregler, and Matthias Nießner.
\newblock {COSMOS}: Catching out-of-context misinformation with self-supervised
  learning, 2021.

\bibitem{mmsys21-challenge-arxiv}
Shivangi Aneja, Cise Midoglu, Duc-Tien Dang-Nguyen, Michael~Alexander Riegler,
  Paal Halvorsen, Matthias Niessner, Balu Adsumilli, and Chris Bregler.
\newblock {MMSys'21} grand challenge on detecting cheapfakes, 2021.

\bibitem{acmmm22-challenge-arxiv}
Shivangi Aneja, Cise Midoglu, Duc-Tien Dang-Nguyen, Sohail~Ahmed Khan, Michael
  Riegler, Pål Halvorsen, Chris Bregler, and Balu Adsumilli.
\newblock {ACM Multimedia} grand challenge on detecting cheapfakes, 2022.

\bibitem{luo2021newsclippings}
G.~Luo, T.~Darrell, and A.~Rohrbach.
\newblock {NewsCLIPpings:} automatic generation of out-of-context multimodal
  media, 2021.

\bibitem{mmsys21-challenge-web}
C.~Midoglu and S.~Aneja.
\newblock Grand challenge on detecting cheapfakes, 2021.

\end{thebibliography}

\end{document}